# Quantum Kernel Learning for Small Dataset Modeling in Semiconductor Fabrication: Application to Ohmic Contact

Zeheng Wang[1,2,†], Fangzhou Wang[3], Liang Li[4], Zirui Wang[5], Timothy van der Laan[2], Ross C. C. Leon[6], Jing-Kai Huang[7], and Muhammad Usman[1,8]

[1] Data61, CSIRO, Clayton, VIC 3168, Australia

[2] Manufacturing, CSIRO, West Lindfield, NSW 2070, Australia

[3] Songshan Lake Materials Laboratory, Dongguan, 523808, China

[4] Academy for Advanced Interdisciplinary Studies, Peking University, Beijing 100871, China

[5] School of Integrated Circuits, Peking University, Beijing 100871, China

[6] Quantum Motion Ltd, London N7 9HJ, United Kingdom

[7] Department of Systems Engineering, City University of Hong Kong, Hong Kong 999077, China

[8] School of Physics, The University of Melbourne, Parkville, VIC 3010, Australia

**† Corresponding author. Email: zenwang@outlook.com**

## Abstract

Modeling complex semiconductor fabrication processes such as Ohmic contact formation remains challenging due to high-dimensional parameter spaces and limited experimental data. While classical machine learning (CML) approaches have been successful in many domains, their performance degrades in small-sample, nonlinear scenarios. In this work, we investigate quantum machine learning (QML) as an alternative, exploiting quantum kernels to capture intricate correlations from compact datasets. Using only 159 experimental GaN HEMT samples, we develop a quantum kernel-aligned regressor (QKAR) combining a shallow Pauli-Z feature map with a trainable quantum kernel alignment (QKA) layer. All models, including seven baseline CML regressors, are evaluated under a unified PCA-based preprocessing pipeline to ensure a fair comparison. QKAR consistently outperforms classical baselines across multiple metrics (MAE, MSE, RMSE), achieving a mean absolute error of 0.338 Ω·mm when validated on experimental data. We further assess noise robustness and generalization through cross-validation and new device fabrication. These findings suggest that carefully constructed QML models could provide predictive advantages in data-constrained semiconductor modeling, offering a foundation for practical deployment on near-term quantum hardware. While challenges remain for both QML and CML, this study demonstrates QML's potential as a complementary approach in complex process modeling tasks.

## Introduction

The rapid advancement of semiconductor technology has led to increasingly complex fabrication processes, where device performance is highly sensitive to numerous process parameters[1–5]. However, collecting large, high-quality experimental datasets to model, understand, and optimize such processes remains challenging due to the cost and time constraints associated with semiconductor manufacturing.

Although classical ML (CML) methods have been extensively explored for enhancing fabrication process modeling[1–7], these limitations pose a significant challenge for CML models, which rely on large datasets to generalize effectively and struggle to capture the intricate nonlinear relationships inherent in semiconductor processes[6,7]. While CML has been extensively explored for device characterization and fabrication modeling, its performance remains limited when applied to small, high-dimensional datasets, often resulting in overfitting and poor generalization to unseen data[8]. Moreover, semiconductor fabrication involves intricate nonlinear relationships among process parameters (e.g., annealing temperature, time, and atmospheric conditions), which further complicates modeling using conventional ML techniques. Addressing these challenges requires a different computational paradigm that can effectively capture high-dimensional correlations while also maintaining robustness in data-scarce environments. We hereby explore quantum computing (QC) algorithms to address both these challenges in semiconductor fabrication.

QC has recently emerged as a promising approach for solving computational problems that are intractable by CML methods[9,10]. In particular, quantum machine learning (QML) has gained significant attention due to its ability to efficiently map classical data into high-dimensional quantum Hilbert spaces, enabling the extraction of subtle patterns even from limited data[11–13]. Unlike CML, QML can leverage quantum kernels, which naturally capture complex feature interactions and offer superior generalization for small datasets[14–19]. These properties make kernel-based QML a compelling candidate for semiconductor modeling, where data is often scarce and process parameters are highly interdependent[9,20–22]. Despite its theoretical advantages, QML has yet to demonstrate improvements over CML when applied to semiconductor fabrication modeling.

In this work, we present what is, to our knowledge, the first application of quantum machine learning (QML) for modeling the formation of Ohmic contacts in GaN high-electron-mobility transistors (HEMTs)—a critical yet challenging step in semiconductor fabrication. Based on 159 experimentally collected samples and data augmentation via a variational autoencoder (VAE), we developed a quantum kernel-aligned regressor (QKAR) incorporating a Pauli-Z feature map and a quantum kernel alignment (QKA) layer, as shown in Fig. 1. Beyond benchmarking QKAR against seven widely used classical machine learning (CML) models, we also performed additional physical experiments for model validation, kernel spectrum diagnostics to assess stability, and noise robustness tests to emulate realistic quantum hardware behavior. Our results show that QKAR consistently outperforms all CML counterparts across multiple evaluation metrics, achieving a mean absolute error (MAE) of 0.338 Ω·mm. These findings demonstrate the potential of QML for effectively handling high-dimensional, small-sample regression tasks in semiconductor domains and point to promising avenues for its deployment in future real-world applications as quantum hardware continues to mature.

## QKAR Modeling

- ***Build QKAR and quantum ablation studies***

The QKAR framework consists of a modular quantum pipeline, as illustrated in Fig. 2(a): a feature map encodes classical data into quantum states; optionally, a QKA layer is inserted to introduce trainable parameters and enhance kernel expressivity; a quantum kernel estimation module evaluates the fidelity between embedded quantum states, and the resulting kernel matrix is processed by a classical support vector regressor (SVR) for prediction tasks.

In this study, we systematically measured three widely used quantum feature maps[23–25],—Pauli-Z, Pauli-ZZ Linear Entangled, and Pauli-ZZ Fully Entangled—each realized on a five-qubit register initialized in the $|00000\rangle$ state (see Fig. 2(c), green blocks). For each feature map, we further considered the effect of integrating a QKA layer (or not), resulting in six possible quantum embedding configurations.

The Pauli-Z feature map consists of Hadamard gates followed by single-qubit Z-rotations, while both variants of Pauli-ZZ incorporate additional entangling gates (with either linear or full connectivity) to capture more complex correlations in the input features. The optional QKA layer (Fig. 2(c), orange block) comprises trainable Rx and Ry rotations, enabling data-dependent kernel adaptation[15].

Ablation results (Fig. 2(b)) indicate that integrating the QKA layer can improve regression accuracy, particularly when combined with the Pauli-Z feature map (i.e., Pauli-Z-QKA), as evidenced by the lowest MAE and highest $R^2$ scores among all tested configurations. Notably, while increasing the entanglement depth in Pauli-based feature maps is often expected to improve representational capacity, our results suggest that deeper embeddings may degrade performance in this case. This effect is likely due to the limited dataset size: more complex quantum circuits can transform the data into quantum states that are more intricate or fragmented in Hilbert space, making task-relevant similarities harder to detect and thus reducing kernel alignment quality[26,27].

These findings suggest that the careful selection and tuning of both feature maps and trainable kernel alignment layers is essential for optimizing quantum kernel models in practical regression tasks involving semiconductor process data. Consequently, the Pauli-Z-QKA architecture was selected as the optimal embedding strategy for QKAR in this study.

## Potential Quantum Advantage

- ***Benchmarking QKAR against CML models***

The benchmarking procedures for the proposed QKAR and seven CML models are schematically illustrated in Fig. 3 (a more detailed description of the data processing pipeline is provided in the Methods section). To align the quantum kernel with the regression targets, a pre-training phase was conducted using only the training data. During this stage, the QKA layer (Fig. 3(a)) was optimized, gradually aligning the quantum kernel to the target kernel, as demonstrated in Fig. 3(b) and (c). Following pre-training, the QKAR was benchmarked against the seven CML models using a 4-fold cross-validation strategy on a sub-dataset. In each fold, the test set was iteratively rotated to cover all samples, ensuring robust and fair comparison across all models.

- ***Potential QKAR advantage***

As shown in Fig. 3(d)-(f), QKAR consistently achieved the lowest average values across all three evaluation metrics (MAE, MSE, RMSE) compared to all other models. The red dashed line represents the expected error when using the training set's mean value for predictions. All models substantially outperformed the baseline predictor, indicating that they effectively learnt meaningful patterns from the data, enabling a fair comparison of their relative performance.

Notably, the results partially indicate that the dataset exhibits a strong linear structure. Classical nonlinear models such as Support Vector Machine (SVM), Decision Tree (DT), Gradient Boosting (GB), XGBoost (XGB), AdaBoost (AB), and Deep Learning (DL) yield the highest prediction errors, suggesting that their expressive power does not translate into performance gains on this task — likely due to the dominance of linear correlations (e.g., higher doping levels typically result in lower Ohmic contact resistance). While linear models like Elastic Net (EN) perform better, their errors remain significantly higher than those of QKAR, implying that conventional methods fail to fully capture subtle feature interactions associated with nonlinear patterns. In contrast, QKAR, by exploiting quantum kernel methods, consistently achieves the lowest MAE, MSE, and RMSE. This suggests its superior capacity to embed both linear and nonlinear patterns into a high-dimensional quantum feature space, where complex interactions — particularly nonlinear ones — become more separable. These findings confirm the advantage of quantum kernel-based models in handling compact datasets that combine strong linear trends with embedded nonlinear complexities, such as those encountered in semiconductor process modeling.

Furthermore, we performed theoretical diagnostics to assess the stability and information structure of the learned quantum kernel. Analyzing the eigenvalue spectrum and the $R^2$ performance with respect to kernel rank (see Supporting Information), we observed that the kernel exhibits a rapidly decaying but non-degenerate spectrum. This indicates a low-rank yet numerically stable structure that captures the essential predictive components while suppressing noise and redundancy.

To evaluate the relative performance of QKAR, we computed its percentage improvement over classical models, as shown in Fig. 3(g)–(i). Under the current small-data regression task, QKAR achieved at least 15.2% improvement in MAE, and no less than 20.1% and 8.8% in MSE and RMSE, respectively. These results indicate that quantum kernel methods can provide measurable predictive benefits in data-scarce modeling scenarios.

To further validate the effectiveness of the proposed QKAR, we also benchmarked it against a Quantum Neural Network (QNN) model (details are shown in SF.2 and SF.3). The QNN model achieved an MAE of 0.54 and an MSE of 0.89. As evident from these results, the QKAR significantly outperforms the QNN model, further reinforcing its superior modeling capability.

## QKAR with Qubit Noise

Noise robustness of QKAR was evaluated, as illustrated in Fig. 4. We adopted the best-performing Pauli-Z-QKA architecture and injected both single-qubit and two-qubit depolarizing noise (DN) into the quantum circuit during simulation (Fig. 4(a)). In this experiment, the two-qubit noise rate was fixed at 0.5% to simulate a stress-test scenario, while the single-qubit noise level was varied. As shown in Fig. 4(b), introducing a 1% single-qubit DN leads to a measurable yet moderate degradation in QKAR's performance. The MAE increases relative to the noiseless case but still remains within the performance window of classical models. While RMSE and $R^2$ decline, the model retains predictive capability, suggesting that QKAR could remain viable for real-world deployment even under relatively high qubit noise levels.

To further assess robustness, we explored an exaggerated noise scenario at 5% DN (Fig. 4(c)). This level exceeds what would typically be encountered in current usable quantum hardware systems. Under this unrealistic condition, model performance deteriorates significantly: the kernel alignment fails to converge, the normalized loss oscillates near zero, and the learned kernel matrix loses meaningful structure. While not representative of physical systems, this test provides insight into QKAR's breakdown threshold under extreme decoherence.

Importantly, most state-of-the-art NISQ platforms exhibit effective depolarization rates below 0.5%[28]. While real quantum hardware involves a broader set of noise mechanisms beyond simple depolarizing channels, many of these effects can be mitigated or corrected using advanced techniques such as circuit-level noise suppression, compilation strategies, and quantum error correction[29,30]. Combined with ongoing progress in these areas, future quantum processors are expected to offer significantly improved fidelity. In this context, the performance of QKAR observed under ideal or low-noise conditions provides a realistic and meaningful preview of its potential utility in near-future applications.

## Verification on Experimental Data

To validate the predictive capability of QKAR, we conducted additional experimental verification, as shown in Fig. 5. The fabrication process is illustrated in Fig. 5(a) and more details can be found in Fig. 5 (b) and Methods. The experiments were performed on two wafers (Fig. 5(c)), each with different material compositions typically used in modern GaN HEMT research. The measured and predicted contact resistances $R_C$, along with the absolute errors, are presented in Fig. 5(d).

Across all test samples, the QKAR predictions exhibited strong agreement with experimental measurements, achieving a MAE of 0.338 Ω·mm. This result is significantly lower than the MAE of other models, reinforcing QKAR's accuracy and robustness in modeling the Ohmic contact process. The consistent performance across different material and process variations highlights the quantum kernel approach's adaptability, demonstrating its potential as a reliable predictive tool for semiconductor manufacturing and research applications.

## Conclusion

In this work, quantum machine learning was applied to semiconductor process modeling, focusing on the Ohmic contact formation in GaN HEMTs. A QKAR was developed and benchmarked against seven classical machine learning models using a unified PCA-reduced dataset consisting of 159 experimental samples. Under four-fold cross-validation, QKAR consistently outperformed the classical baselines across multiple evaluation metrics, demonstrating robust predictive capability in the small-data regime.

To further validate the model's generalization ability, we fabricated five additional test devices under varied process conditions. QKAR achieved a mean absolute error of 0.338 Ω·mm when predicting the contact resistances of these unseen samples, confirming its practical potential in extrapolating to new fabrication settings.

All classical and quantum models were evaluated under the same feature representations and validation protocols to ensure fairness. While QKAR performed favorably in this specific context, classical models could potentially achieve improved results with more extensive hyperparameter tuning or alternative architectures. Additionally, although this study was conducted on classical quantum simulators, the quantum circuits used are compatible with current NISQ hardware. As quantum processors improve in fidelity and scale, the deployment of QML models in real semiconductor workflows may become increasingly viable. This work thus represents a foundational step toward integrating quantum learning techniques into data-driven fabrication modeling.

## Methods

- ***Dataset of GaN Ohmic contacts***

GaN HEMTs are widely used in high-frequency, high-power, and high-efficiency electronic applications due to their wide bandgap, high electron mobility, and superior thermal stability[31–36]. Compared to traditional silicon-based transistors, GaN HEMTs offer lower conduction losses and higher breakdown voltages, making them ideal for power electronics, RF amplifiers, and next-generation communication systems. However, the fabrication of Ohmic contacts on GaN HEMTs remains a critical challenge, as the formation process involves complex interactions between metal stacks, annealing conditions, and the wide-bandgap AlGaN barrier layer, etc., all of which influence the contact resistance and overall device performance[37,38]. Given the highly sensitive and nonlinear nature of these process parameters, GaN HEMT data serves as an excellent test case for evaluating advanced modeling techniques, particularly in scenarios where data collection is limited due to the cost and complexity of fabrication.

To form the Ohmic contact on GaN HEMT, a metal stack is typically used as the electrode at the AlGaN surface, and an annealing process is required to enhance the electrical contact between the metal electrode and the conducting two-dimensional electron gas at the AlGaN/GaN interface. Therefore, the data of Al content, AlGaN thickness, metal stack type, and annealing conditions were recorded.

- ***Data pipeline and modeling environment***

We extracted data on Ohmic contacts from 159 GaN HEMT devices reported in Ref.[6] and encoded each sample (i.e., each fabrication recipe) into 37 one-hot–encoded features plus the contact resistance label (contact resistance, $R_C$).

To reduce the dimensionality of the original 37-feature dataset, we first applied a PCA-based method to the entire dataset before any data splitting. The resulting five-dimensional representations were then used throughout all subsequent modeling stages. For VAE-based data augmentation, QKAR ablation studies, and noise robustness testing, we performed multiple randomized 80%/20% splits of the 5D dataset to rapidly train and evaluate models, with results averaged across repetitions to ensure statistical reliability. The final augmented training set comprised 381 samples (including both experimental and VAE-synthesized data), while the test sets used for each split consisted solely of experimental data. For benchmarking against CML models, we adopted 4-fold cross-validation on the 5D dataset to ensure a robust and fair performance comparison across all models. See Fig. 6 for a straightforward understanding.

CML baselines were implemented in Python 3.12 using Scikit-learn 1.5.1 following the configurations of Ref.[39], and the QKAR algorithm was coded in Qiskit 1.2.4 with its Machine Learning extension 0.8.2. All CML models' key configurations can be found in Supporting Information.

Fig. 6 illustrates the overall data-processing pipeline: Phase II ("quantum ablation") uses five random 80%/20% splits to identify the optimal QKAR architecture rapidly, while Phase III ("benchmarking") employs 4-fold cross-validation on the post-PCA five-dimensional dataset to compare QKAR against seven CML models.

The detailed QKAR construction workflow is shown in Fig. 7: after VAE augmentation (Fig. 7(a)), quantum circuits are instantiated from the components in Fig. 7(b) to produce the embedding (Fig. 7(c)), qubit states are visualized for a sample input (Fig. 7(d)), the kernel matrix is computed (Fig. 7(e)), and finally kernel alignment and SVR training are performed before test-set evaluation (Fig. 7(f)).

While we have made efforts to ensure a fair comparison across all models by applying consistent preprocessing and validation procedures, it is important to note that the hyperparameters of CML models in this study were selected based on established configurations from prior work rather than exhaustive tuning. Likewise, the quantum kernel SVR was used with default parameters without additional optimization. We therefore acknowledge that the performance of classical and quantum models could potentially be improved further with systematic hyperparameter search or alternative optimization strategies. A more comprehensive tuning procedure will be explored in future work to fully assess the comparative modeling capacity of QML and CML frameworks.

- ***QKAR model***

All QML experiments were conducted on a classical simulator (Qiskit Aer) running on local hardware. While real quantum hardware was not used, all circuits were NISQ-compatible and designed with potential hardware deployment in mind.

The QKAR model proposed in this study is built upon static quantum feature maps and a variational QKA layer. This design leverages the high-dimensional Hilbert space of quantum states to improve regression performance. In contrast to the self-adaptive QKAR architecture introduced in Ref. [15] and other dynamic kernel alignment approaches described in Refs.[40–42], the QKAR developed here adopts a simplified quantum embedding scheme, thereby avoiding the substantial computational overhead associated with deep variational quantum circuits.

Specifically, the model employs a shallow, one-layer Pauli-Z feature map, which offers sufficient expressivity while maintaining computational efficiency, examined by the quantum ablation study. The subsequent simple QKA layer promotes kernel stability, enabling well-aligned quantum kernels that can be precomputed and reused with minimal pre-training. This architecture is particularly well suited to scenarios involving small datasets with mixed linear–nonlinear structure, where quantum embeddings are beneficial but do not necessitate complex, continuously adaptive circuits — a setting that aligns closely with real-world NISQ-era applications.

To further evaluate the robustness and internal behavior of QKAR, we conducted comprehensive benchmarking against multiple CML baselines under varied semiconductor process conditions. In addition to external performance comparisons, we performed a spectral analysis of the learned quantum kernel matrices. By examining the eigenvalue distributions and the predictive performance as a function of kernel rank (see Supporting Information), we verified that the kernel retains a non-degenerate structure and captures meaningful task-specific features. This analysis confirms that QKAR maintains numerical stability and avoids kernel collapse, even in the small-sample regime.

The core mathematical framework consists of feature mapping, variational QKA, quantum kernel calculation, and regression optimization.

Given an input data point $\boldsymbol{x} \in \mathbb{R}^n$, a quantum feature map $\boldsymbol{\Phi}(\boldsymbol{x})$ is used to encode it into a quantum state:

$$|\Phi(\boldsymbol{x})\rangle = U_F(\boldsymbol{x})|0\rangle^{\otimes n} \tag{1}$$

Where $U_F(x)$ is a parameterized unitary transformation defined by the chosen feature map and $n$ is the dimension of the circuit where we have $n = 5$ in this work. In this work, we adopted the 1-level Pauli-Z feature map, which applies Hadamard gates and Pauli rotations, as shown in Fig. 7. The transformation can be expressed as:

$$U_F(\boldsymbol{x}) = \bigotimes_{j=1}^{n} \left(R_z(x_j) \cdot H_j\right) \tag{2}$$

where $H$ represents Hadamard gates and $R_z(x_j) = \exp\left(-i\frac{x_j}{2}Z\right)$ represents parameterized Pauli rotations encoding input features.

The feature map is followed by a parametric TwoLocal QKA layer composed of alternating $R_y$ and $R_x$ gates with interleaved entanglement layers. The transformation can be written as:

$$U_{QKA}(x) = \left[CNOT \cdot \bigotimes_{j=1}^{n} R_y(\theta_j)R_x(x_j)\right]^r \tag{3}$$

where $R_x(x_j) = \exp\left(-i\frac{x_j}{2}X\right)$ encodes classical input features into rotation angles, and $R_y(\theta_j) = \exp\left(-i\frac{\theta_j}{2}Y\right)$ adds additional expressivity through fixed or trainable parameters $\theta_j$. The entanglement layer $CNOT$ is implemented using $CNOT$ gates in a linear topology (see Fig. 7 (c)). The number of repetitions $r = 1$ controls the depth of the circuit.

The similarity between two data points $x_i$ and $x_j$ in the quantum feature space is computed using the quantum kernel function:

$$K(x_i, x_j) = |\langle U(x_i) | U(x_j) \rangle|^2 \qquad (4)$$

which measures the fidelity between the quantum states corresponding to the embedded sample pairs, e.g., $U(x_i)$ and $U(x_j)$. The quantum kernel matrix $K$, with elements $K_{ij} = K(x_i, x_j)$, is then used as input for regression.

A SVR is used to perform the regression task in classical space. Given the training set $\{(x_i, x_i)\}_{i=1}^{m}$ with input-output pairs, the regression function is obtained by solving the following optimization problem:

$$\min_{\alpha} \frac{1}{2} \sum_{i,j} \alpha_i \alpha_j K(x_i, x_j) - \sum_i \alpha_i y_i \qquad (5)$$

subject to the constraints:

$$0 \le \alpha_i \le C, \quad \sum_i \alpha_i = 0 \qquad (6)$$

where $C$ is a regularization parameter. The predicted output for a new input $x$ is then given by:

$$\hat{y} = \sum_i \alpha_i K(x, x_i) + b \qquad (7)$$

where $b$ is the bias term learned during training.

By utilizing quantum-enhanced feature mapping, fidelity-based kernels, and SVR optimization, QKAR effectively captures nonlinear relationships in the semiconductor process data. The Pauli-Z-QKA quantum embedding further enhances expressibility and entanglement, leading to higher predictive accuracy compared to classical models.

- ***Noise channel***

To simulate realistic quantum execution conditions, we incorporated both single-qubit and two-qubit depolarizing noise channels into the quantum circuit. Specifically, after each single-qubit gate, a single-qubit depolarization channel $\epsilon(\rho)$ with depolarizing probability $p$ was applied. Likewise, a fixed two-qubit depolarizing noise with a 0.5% probability was applied after each two-qubit gate. This noise model transforms a qubit's density matrix $\rho$ as:

$$\epsilon(\rho) = (1 - p)\rho + \frac{p}{3}(X\rho X + Y\rho Y + Z\rho Z) \quad (8)$$

where $X, Y, Z$ are the standard Pauli operators, and $p$ controls the noise strength. This channel models random gate errors commonly observed in near-term NISQ devices. The inclusion of depolarizing noise allows us to assess the robustness of the quantum kernel method under realistic hardware constraints. The findings suggest that multiple emerging hardware platforms are compatible with the implementation of the proposed QKAR[43–45].

- ***Fabrication and measurement of GaN Ohmics***

The experimental verification was conducted using two wafers with different AlGaN barrier thicknesses (13 nm and 15 nm) and Al compositions (0.25 and 0.20). Different metal stacks were employed: Ti/Al/Ni/Au and Ti/Al/Ti/TiN. To assess the impact of process conditions, samples underwent distinct annealing treatments: Wafer 1 (Samples 1–3) were annealed at 830°C, 850°C, and 870°C for 30 s, while Wafer 2 (Samples 4–5) were annealed at 500°C and 650°C for 90 s. The fabrication recipes are also listed in Fig. 5(b). All $R_C$ measurements were conducted using a probe station with a Keysight B1500 semiconductor parameter analyzer, based on the standard transmission line model (TLM). The results confirm that QKAR effectively captures the complex relationships between process conditions and electrical properties, providing a highly accurate, data-driven approach to modeling the Ohmic contact formation in GaN HEMTs.

## Acknowledgement

This work was supported in part by the CSIRO Impossible Without You Program. The manuscript was proofread and linguistically improved with the assistance of large language models.

## References

1. Jeong, C. *et al.* Bridging TCAD and AI: Its Application to Semiconductor Design. *IEEE Trans. Electron Devices* **68**, 5364–5371 (2021).
2. Wu, T.-L. & Kutub, S. B. Machine Learning-Based Statistical Approach to Analyze Process Dependencies on Threshold Voltage in Recessed Gate AlGaN/GaN MIS-HEMTs. *IEEE Trans. Electron Devices* **67**, 5448–5453 (2020).
3. Mehta, K. *et al.* Improvement of TCAD Augmented Machine Learning Using Autoencoder for Semiconductor Variation Identification and Inverse Design. *IEEE Access* **8**, 143519–143529 (2020).
4. Lu, A. *et al.* Vertical GaN diode BV maximization through rapid TCAD simulation and ML-enabled surrogate model. *Solid-State Electron.* **198**, 108468 (2022).
5. Singhal, A., Machhiwar, Y., Kumar, S., Pahwa, G. & Agarwal, H. ANN-based framework for modeling process induced variation using BSIM-CMG unified model. *Solid-State Electron.* **220**, 108988 (2024).
6. Wang, Z. *et al.* Improving Semiconductor Device Modeling for Electronic Design Automation by Machine Learning Techniques. *IEEE Trans. Electron Devices* 1–9 (2023) doi:10.1109/TED.2023.3307051.
7. Wang, Z., Li, L. & Yao, Y. A machine learning-assisted model for GaN ohmic contacts regarding the fabrication processes. *IEEE Trans. Electron Devices* **68**, 2212–2219 (2021).
8. Bishop, C. M. *Pattern Recognition and Machine Learning*. (Springer New York, New York, NY, 2016).
9. Biamonte, J. *et al.* Quantum machine learning. *Nature* **549**, 195–202 (2017).
10. Cerezo, M., Verdon, G., Huang, H.-Y., Cincio, L. & Coles, P. J. Challenges and opportunities in quantum machine learning. *Nat. Comput. Sci.* **2**, 567–576 (2022).
11. Caro, M. C. *et al.* Generalization in quantum machine learning from few training data. *Nat. Commun.* **13**, 4919 (2022).
12. Otgonbaatar, S., Schwarz, G., Datcu, M. & Kranzlmüller, D. Quantum Transfer Learning for Real-World, Small, and High-Dimensional Remotely Sensed Datasets. *IEEE J. Sel. Top. Appl. Earth Obs. Remote Sens.* **16**, 9223–9230 (2023).
13. Yung, C. & Usman, M. Clustering by Contour Coreset and Variational Quantum Eigensolver. *Adv. Quantum Technol.* **7**, 2300450 (2024).
14. Glick, J. R. *et al.* Covariant quantum kernels for data with group structure. *Nat. Phys.* (2024) doi:10.1038/s41567-023-02340-9.
15. Wang, Z., Van Der Laan, T. & Usman, M. Self-Adaptive Quantum Kernel Principal Component Analysis for Compact Readout of Chemiresistive Sensor Arrays. *Adv. Sci.* 2411573 (2025) doi:10.1002/advs.202411573.
16. Schuld, M. Supervised quantum machine learning models are kernel methods. Preprint at https://doi.org/10.48550/ARXIV.2101.11020 (2021).
17. Tomono, T. & Tsujimura, K. Quantum kernel learning Model constructed with small data. Preprint at https://doi.org/10.48550/ARXIV.2412.00783 (2024).
18. Tscharke, K., Issel, S. & Debus, P. QUACK: Quantum Aligned Centroid Kernel. in *2024 IEEE*

*International Conference on Quantum Computing and Engineering (QCE)* 1425–1435 (IEEE, Montreal, QC, Canada, 2024). doi:10.1109/QCE60285.2024.00169.

19. Dowling, N. *et al.* Adversarial Robustness Guarantees for Quantum Classifiers. Preprint at https://doi.org/10.48550/ARXIV.2405.10360 (2024).

20. West, M. T. *et al.* Towards quantum enhanced adversarial robustness in machine learning. *Nat. Mach. Intell.* **5**, 581–589 (2023).

21. Havlíček, V. *et al.* Supervised learning with quantum-enhanced feature spaces. *Nature* **567**, 209–212 (2019).

22. Jadhav, A., Rasool, A. & Gyanchandani, M. Quantum Machine Learning: Scope for real-world problems. *Procedia Comput. Sci.* **218**, 2612–2625 (2023).

23. Aksoy, G., Cattan, G., Chakraborty, S. & Karabatak, M. Quantum Machine-Based Decision Support System for the Detection of Schizophrenia from EEG Records. *J. Med. Syst.* **48**, 29 (2024).

24. Daspal, A. Effect of Repetitions and Entanglement on Performance of Pauli Feature Map. in *2023 IEEE International Conference on Quantum Computing and Engineering (QCE)* 278–279 (IEEE, Bellevue, WA, USA, 2023). doi:10.1109/QCE57702.2023.10241.

25. Creevey, F. M., Heredge, J. A., Sevior, M. E. & Hollenberg, L. C. L. Kernel Alignment for Quantum Support Vector Machines Using Genetic Algorithms. Preprint at https://doi.org/10.48550/ARXIV.2312.01562 (2023).

26. Larocca, M. *et al.* Barren plateaus in variational quantum computing. *Nat. Rev. Phys.* **7**, 174–189 (2025).

27. Tomasi, J., Anthoine, S. & Kadri, H. Benign Overfitting with Quantum Kernels. Preprint at https://doi.org/10.48550/arXiv.2503.17020 (2025).

28. Yang, C. H. *et al.* Silicon qubit fidelities approaching incoherent noise limits via pulse engineering. *Nat. Electron.* **2**, 151–158 (2019).

29. Bravyi, S. *et al.* High-threshold and low-overhead fault-tolerant quantum memory. *Nature* **627**, 778–782 (2024).

30. Cai, Z. *et al.* Quantum error mitigation. *Rev. Mod. Phys.* **95**, 045005 (2023).

31. Wang, Z. *et al.* A high-performance tunable LED-compatible current regulator using an integrated voltage nanosensor. *IEEE Trans. Electron Devices* **66**, 1917–1923 (2019).

32. Li, C., Chen, X. & Wang, Z. Review of the AlGaN/GaN High-Electron-Mobility Transistor-Based Biosensors: Structure, Mechanisms, and Applications. *Micromachines* **15**, 330 (2024).

33. Wang, Z., Wang, F., Guo, S. & Wang, Z. Simulation study of high-reverse blocking AlGaN/GaN power rectifier with an integrated lateral composite buffer diode. *Micro Nano Lett.* **12**, 660–663 (2017).

34. Wang, Z. Proposal of a novel recess-free enhancement-mode AlGaN/GaN HEMT with field-assembled structure: a simulation study. *J. Comput. Electron.* **18**, 1251–1258 (2019).

35. Wang, Z. *et al.* Simulation study on AlGaN/GaN diode with Γ-shaped anode for ultra-low turn-on voltage. *Superlattices Microstruct.* **117**, 330–335 (2018).

36. Wang, Z. *et al.* A High-Performance Tunable LED-Compatible Current Regulator Using an Integrated Voltage Nanosensor. *IEEE Trans. Electron Devices* **66**, 1917–1923 (2019).

37. He, J. *et al.* Recent Advances in GaN-Based Power HEMT Devices. *Adv. Electron. Mater.* **7**, 2001045 (2021).

38. Liu, A.-C. *et al.* The Evolution of Manufacturing

Technology for GaN Electronic Devices. *Micromachines* **12**, 737 (2021).

39. Wang, Y., Li, C. & Wang, Z. Advancing Precision Medicine: VAE Enhanced Predictions of Pancreatic Cancer Patient Survival in Local Hospital. *IEEE Access* **12**, 3428–3436 (2024).

40. Hubregtsen, T. *et al.* Training quantum embedding kernels on near-term quantum computers. *Phys. Rev. A* **106**, 042431 (2022).

41. Meyer, J. J. *et al.* Exploiting Symmetry in Variational Quantum Machine Learning. *PRX Quantum* **4**, 010328 (2023).

42. Incudini, M. *et al.* Automatic and Effective Discovery of Quantum Kernels. *IEEE Trans. Emerg. Top. Comput. Intell.* 1–10 (2024) doi:10.1109/TETCI.2024.3499993.

43. Steinacker, P. *et al.* A 300 mm foundry silicon spin qubit unit cell exceeding 99% fidelity in all operations. Preprint at https://doi.org/10.48550/arXiv.2410.15590 (2024).

44. Wang, Z. *et al.* Jellybean Quantum Dots in Silicon for Qubit Coupling and On-Chip Quantum Chemistry. *Adv. Mater.* **35**, 2208557 (2023).

45. Reiner, J. *et al.* High-fidelity initialization and control of electron and nuclear spins in a four-qubit register. *Nat. Nanotechnol.* **19**, 605–611 (2024).

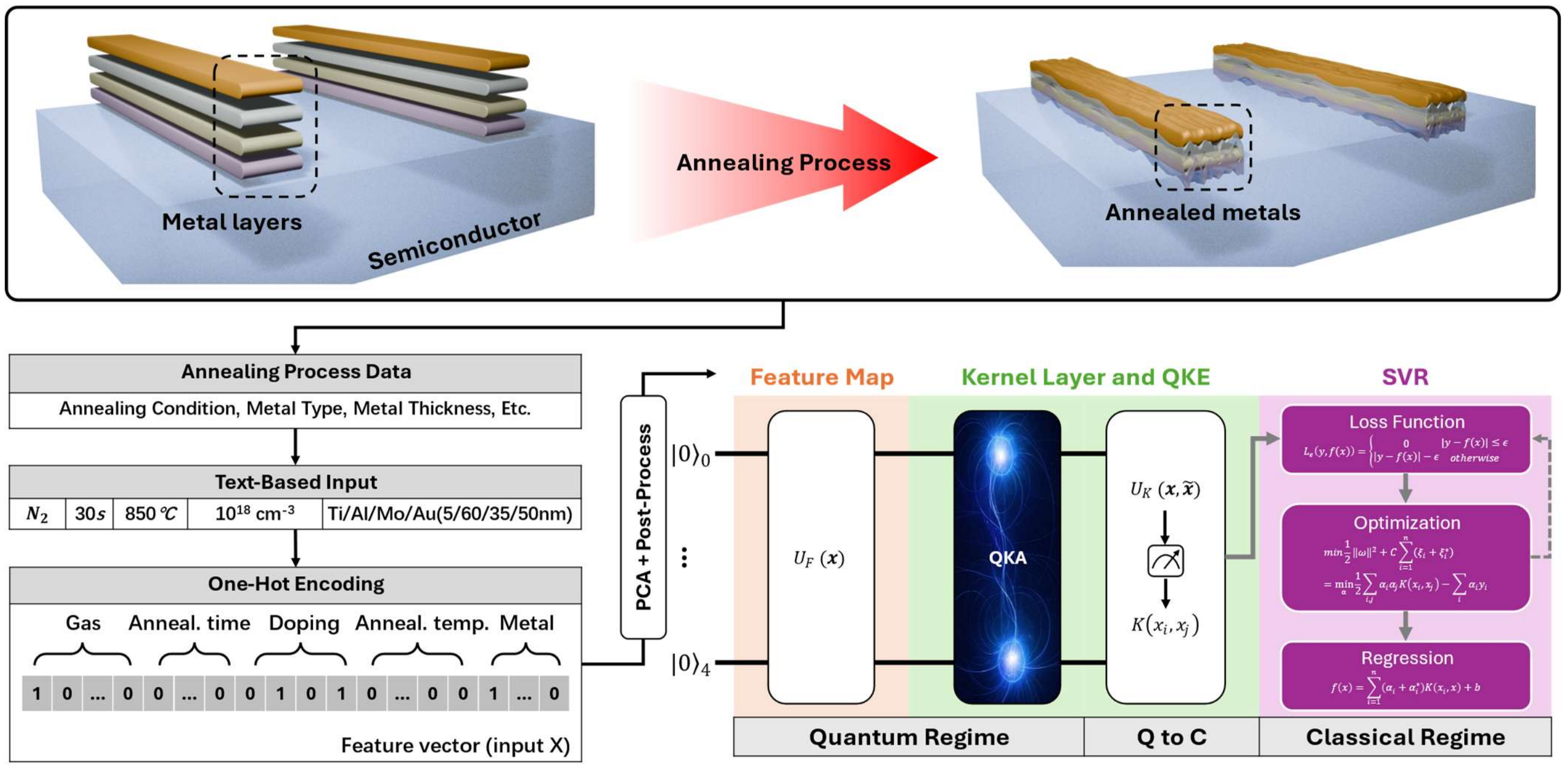

Fig. 1. Schematic representation of the quantum machine learning-based modeling process for the Ohmic contact formation in GaN HEMTs. The process begins with data extraction and preparation, where key fabrication parameters such as annealing conditions, metal composition, and thickness are collected. The raw data is then encoded into a dataset by one-hot encoding, followed by a PCA-based dimensional reduction process (37D to 5D). A VAE-based data augmentation is applied to enhance the dataset using the training set only. The modeling framework consists of a hybrid quantum-classical pipeline. In the Quantum Regime, a quantum feature map and a quantum kernel alignment layer encode input data into quantum states, followed by a quantum kernel layer that computes pairwise similarities (i.e., fidelity kernels) in a high-dimensional Hilbert space. The Quantum-to-Classical (Q to C) transition transfers the computed quantum kernel matrix to classical space, where it serves as input for the Classical Regime. Here, a support vector regressor (SVR) processes the kernel data to train a predictive model for Ohmic contact resistance, leveraging QML's enhanced ability to capture high-dimensional correlations from small datasets. $L_\epsilon(y, f(x))$ is the $\epsilon$-insensitive loss function that measures the error between the target $y$ and the predicted value $f(x)$; $\omega$ is the weights vector of the model, and $C$ is the regularization parameter that balances model complexity and training error. $\xi_i$ and $\xi_i^*$ are slack variables that allow for deviations from the $\epsilon$-insensitive zone. $\alpha_i$ and $\alpha_i^*$ are Lagrange multipliers used in the dual optimization problem. $K(x_i, x)$ is the kernel function (by quantum states fidelity) that maps input data $x$ into a higher-dimensional space to capture non-linear relationships.

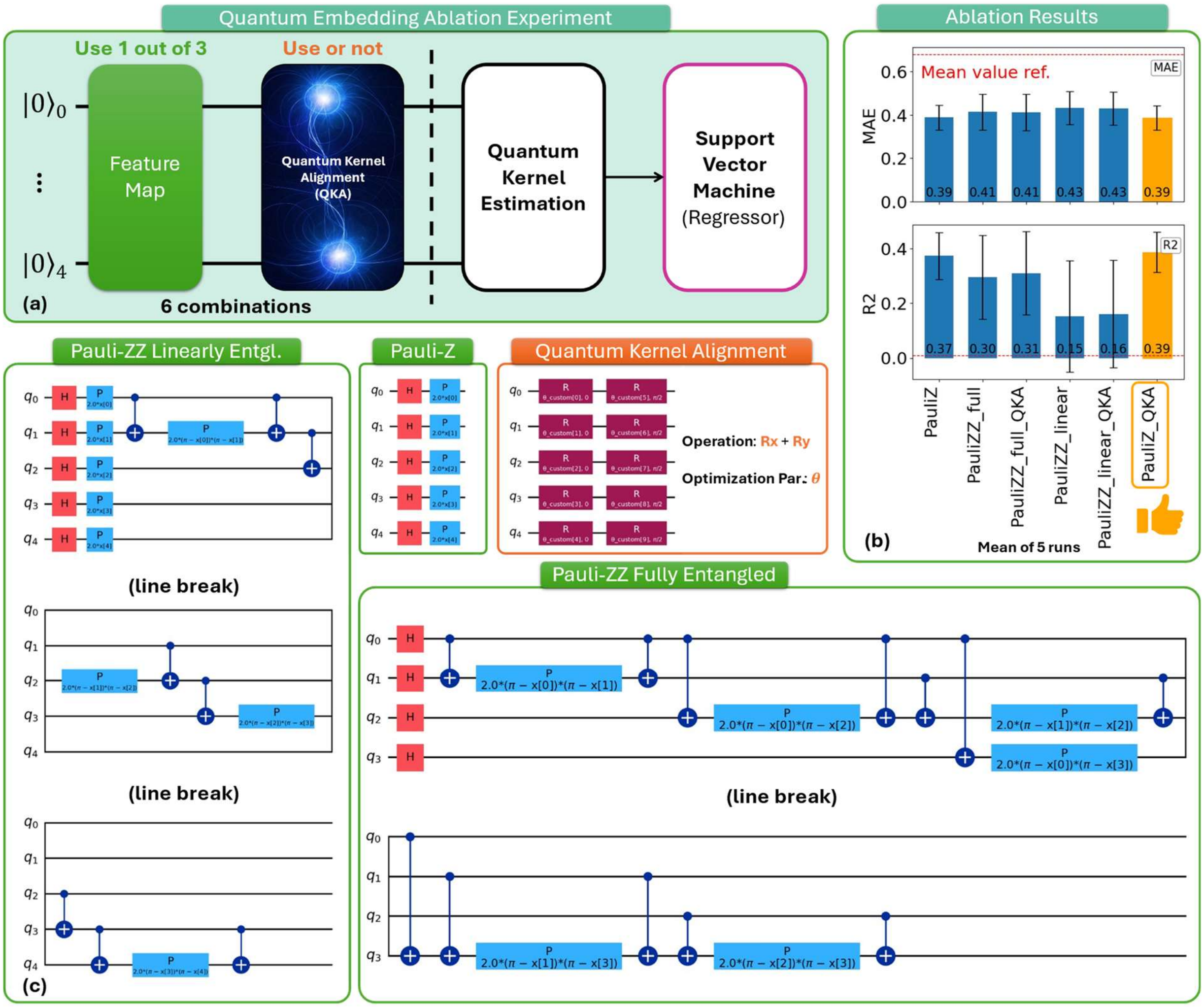

Fig. 2. Quantum ablation study for optimizing the performance of QKAR. (a) Three mainstream quantum feature maps are investigated—Pauli-Z, linearly entangled Pauli-ZZ, and fully entangled Pauli-ZZ—to evaluate their influence on the expressibility and generalization capability of the QKAR model. The effect of the QKA layer, used for kernel alignment, is also ablated to assess its contribution. In total, six combinations of quantum embeddings are compared. (b) Performance metrics (MAE, MSE, RMSE) for the six configurations reveal the impact of different embedding strategies on regression accuracy. (c) Quantum circuit structures of the feature map and QKA components used in each configuration of this study.

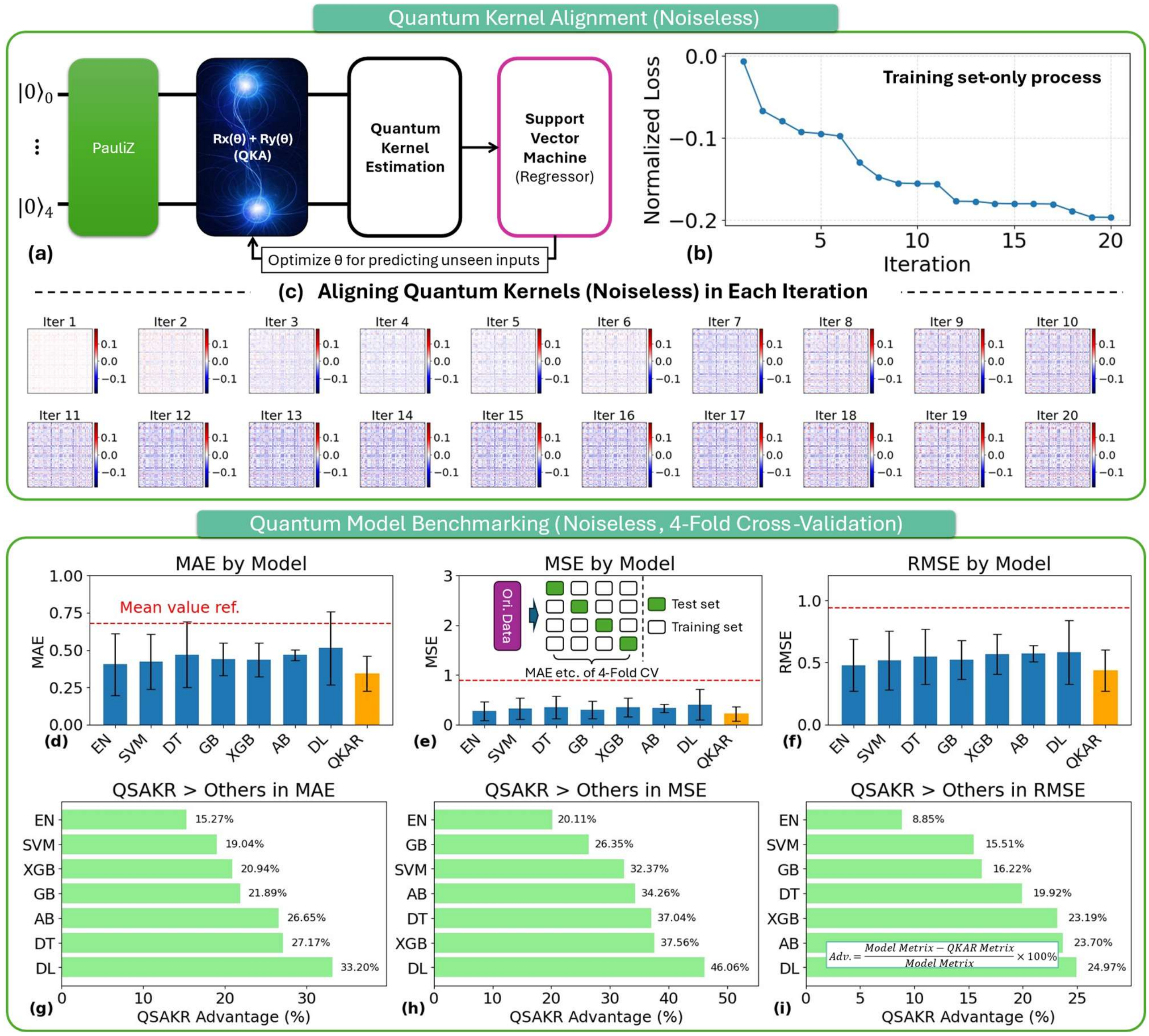

Fig. 3. Training of the kernel-aligned, noiseless QKAR and its benchmarking against CML models. The selected QKAR architecture combines a Pauli-Z feature map with a QKA layer, where the training set is used to optimize the alignment parameter $\theta$. Panel (b) shows that the normalized training loss converges within 20 iterations, while panel (c) illustrates the evolution of kernel differences during alignment. Model performance is evaluated using 4-fold cross-validation. Panels (d)–(f) compare the MAE, MSE, and RMSE of QKAR with various CML models. Panels (g)–(i) quantify the potential quantum advantage of QKAR over classical baselines, with advantage percentages computed as defined in the equation shown in the inset of panel (i).

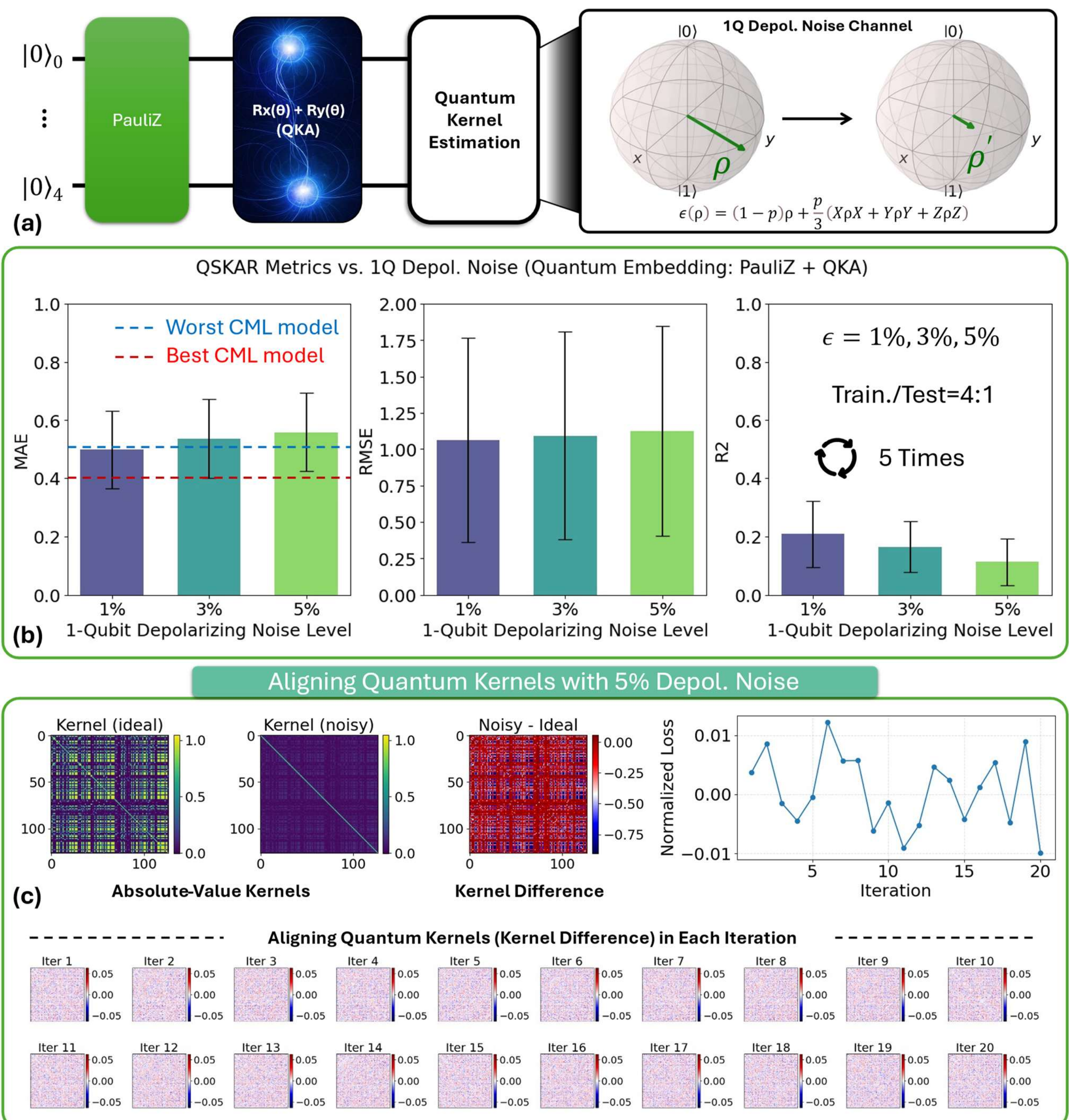

Fig. 4. Evaluating the performance of QKAR under varying levels of single-qubit depolarizing noise. (a) Schematic of the QKAR architecture and the applied depolarizing noise channel. (b) Evaluation metrics (MAE, MSE, RMSE) of the QKAR across different noise levels, with a 4:1 train-test split and results averaged over five independent runs. (c) Kernel alignment process under 5% depolarizing noise, where the kernel fails to converge and the normalized loss fluctuates near zero, indicating ineffective optimization.

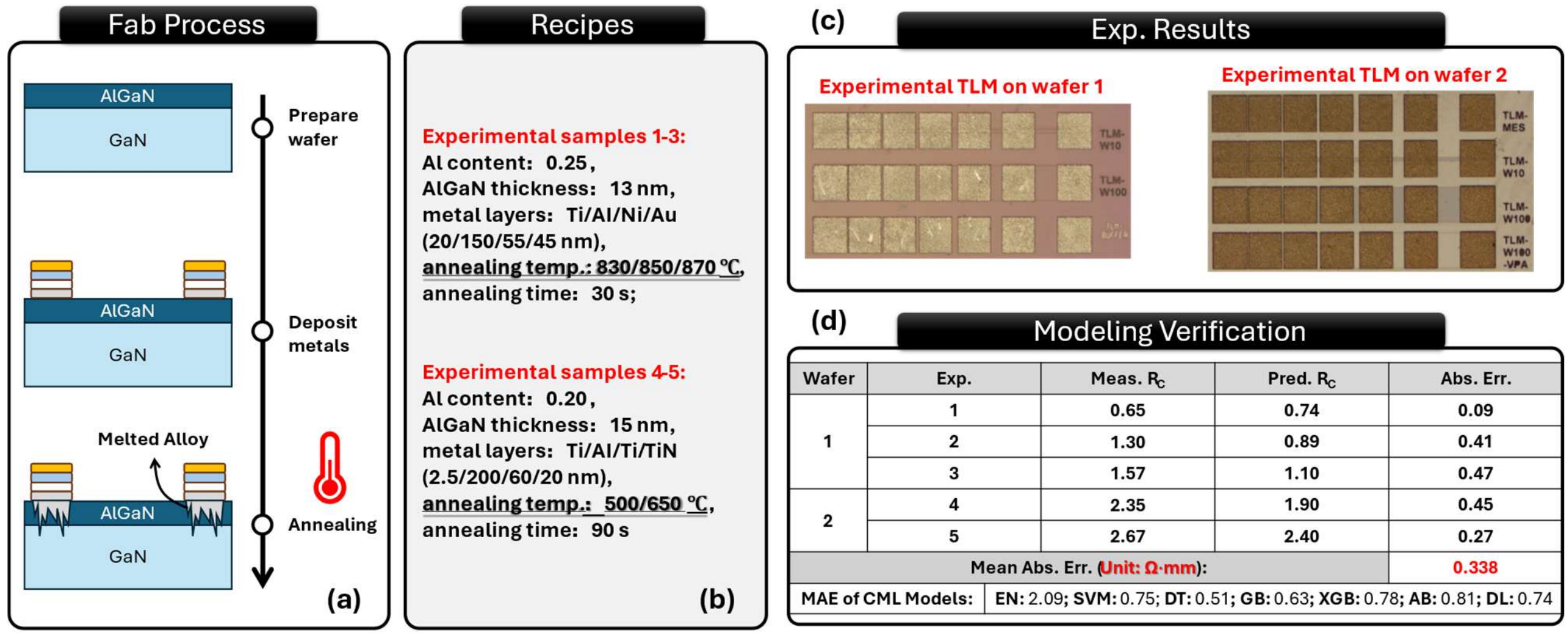

| Wafer | Exp. | Meas. $R_C$ | Pred. $R_C$ | Abs. Err. |
|---|---|---|---|---|
| 1 | 1 | 0.65 | 0.74 | 0.09 |
| | 2 | 1.30 | 0.89 | 0.41 |
| | 3 | 1.57 | 1.10 | 0.47 |
| 2 | 4 | 2.35 | 1.90 | 0.45 |
| | 5 | 2.67 | 2.40 | 0.27 |
| Mean Abs. Err. (Unit: Ω·mm): | | | | 0.338 |
| MAE of CML Models: | EN: 2.09; SVM: 0.75; DT: 0.51; GB: 0.63; XGB: 0.78; AB: 0.81; DL: 0.74 | | | |

Fig. 5. Experimental verification of the proposed QKAR model. (a) and (b) show the experimental setup, in which two wafers with distinct AlGaN compositions, barrier thicknesses, and metal stacks were processed to form Ohmic contacts under varied annealing conditions. (c) displays an optical microscopy image of the fabricated devices. (d) summarizes the measured and predicted contact resistance values for each sample, along with the corresponding absolute errors. The QKR model achieves a MAE of 0.338 Ω·mm, substantially outperforming classical models. These results validate the effectiveness of QKR in modeling high-dimensional, small-sample semiconductor fabrication processes.

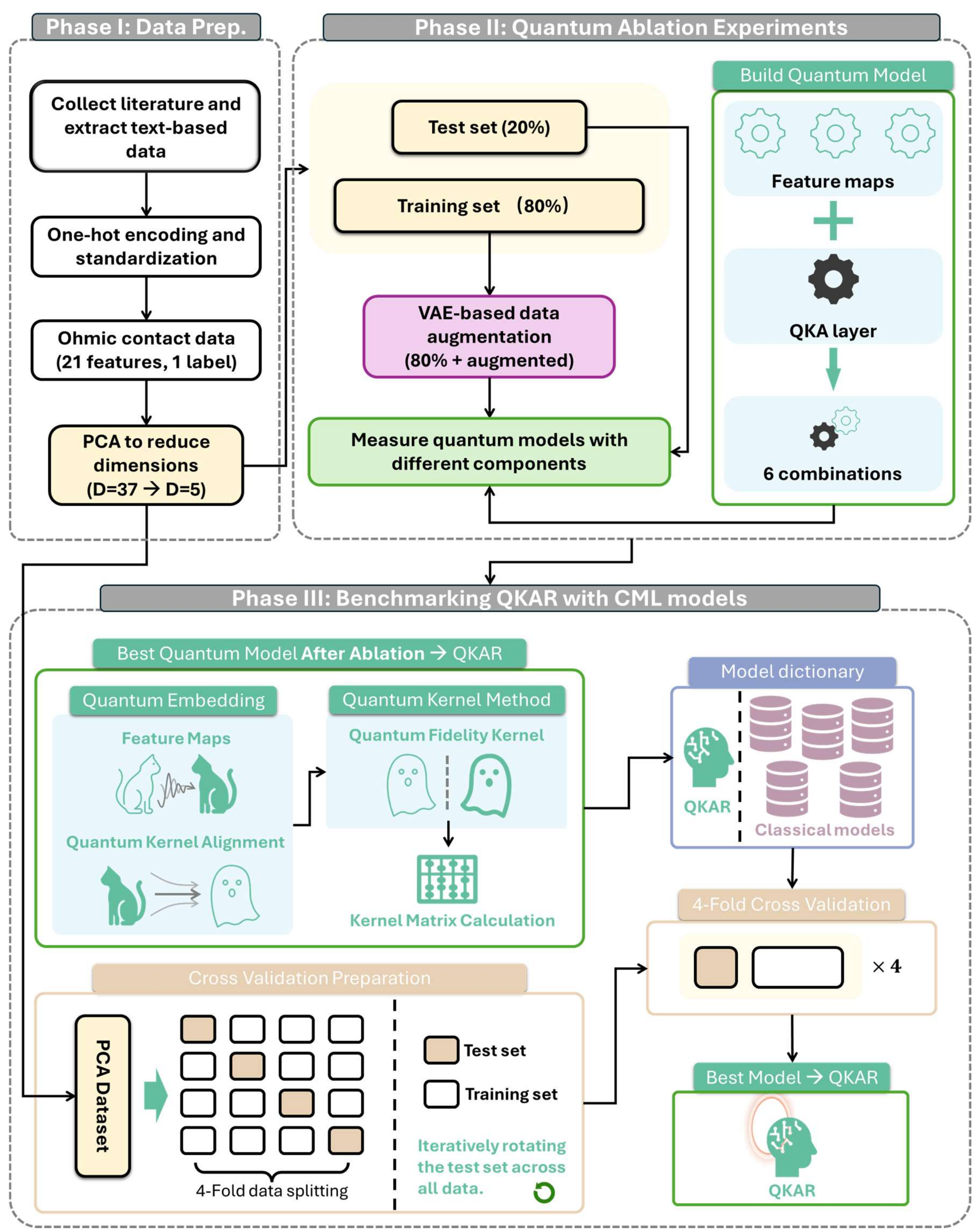

Fig. 6. Data processing pipeline used in this study. In Phase I, raw data were collected and one-hot encoded, followed by PCA-based dimensional reduction method for compressing the features from 37 to 5 dimensions. In Phase II, the resulting 5D dataset was randomly split into training (80%) and test (20%) sets across five repetitions for quantum ablation studies to identify the optimal QKAR architecture. The selected QKAR was then benchmarked against seven CML models using 4-fold cross-validation on a sub-dataset, repeated four times. All data underwent identical preprocessing before being input into the QML and CML models to ensure fair comparison

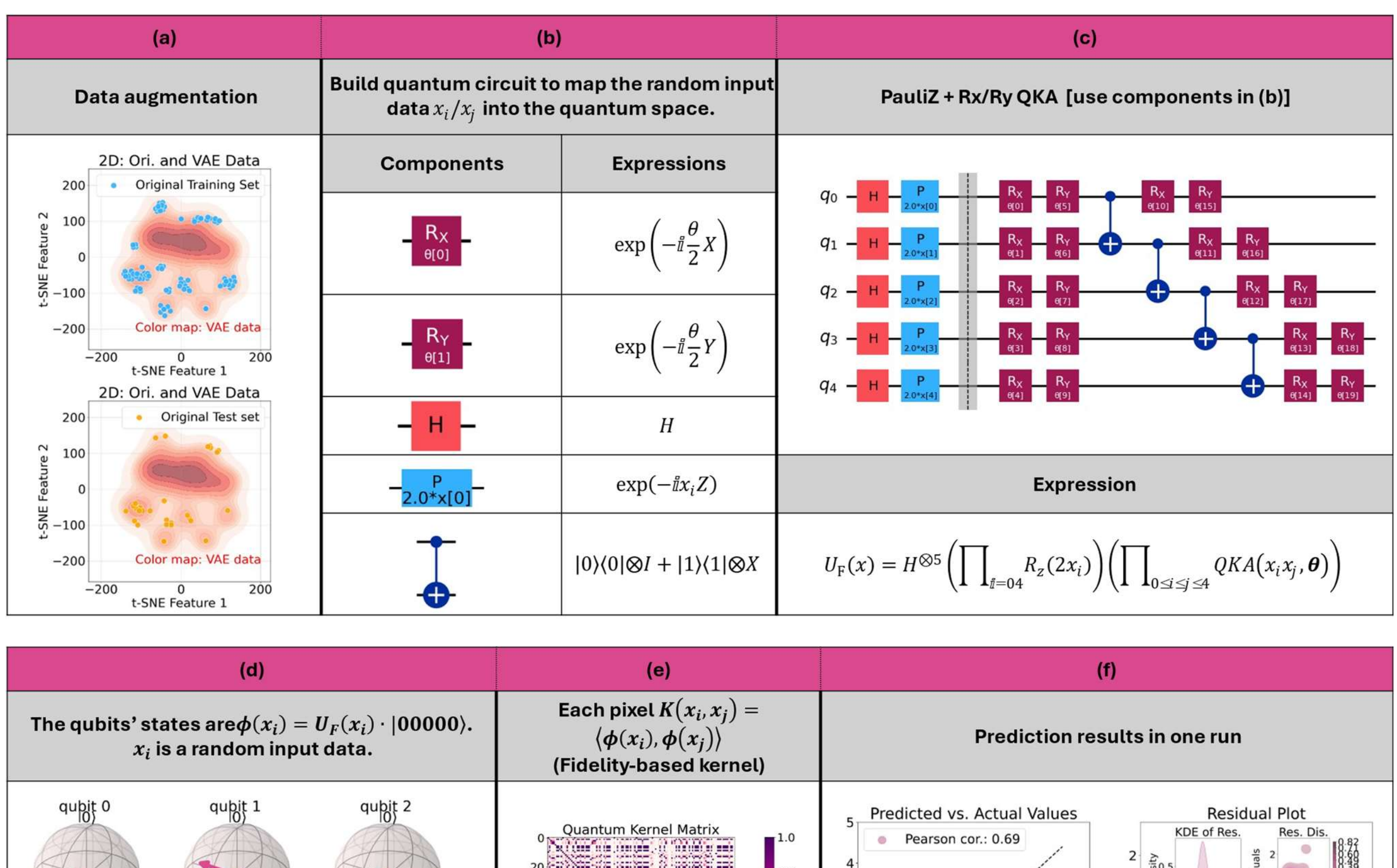

Fig. 7. The process of how to build the QML model. (a) The VAE-based technique augmented the data for training the model; (b) and (c) The quantum gates and the circuit for the feature mapping; (d) the qubits' expectation and the phases shown in the Bloch sphere when inputting random data $x_i$ after mapping; (e) the quantum kernel used for regression, where meaningful correlations between device samples (each corresponding to a unique fabrication recipe) can be observed, suggesting the possible patterns of the resistance associated with the recipes; (f)-(i) the benchmarking of the modeling results: the correlation is strong and most residuals are small around 0.